\documentclass[runningheads]{llncs}
\usepackage[T1]{fontenc}
\usepackage{graphicx}
\usepackage{booktabs}
\usepackage{dirtytalk}
\usepackage{csquotes}

\usepackage{enumitem}
\setlist[itemize]{itemsep=0.3em}

\usepackage{xcolor}
\usepackage{array}

\usepackage{hyperref}
\usepackage[normalem]{ulem}

\begin{document}
\title{BR-TaxQA-R: A Dataset for Question Answering with References for Brazilian Personal Income Tax Law, including case law}

\titlerunning{BR-TaxQA-R}
%
\author{Juvenal Domingos Júnior\inst{1} \and
Augusto Faria\inst{1} \and
E. Seiti de Oliveira\inst{1} \and
Erick de Brito\inst{2} \and
Matheus Teotonio\inst{2} \and
Andre Assumpção\inst{3} \and
Diedre Carmo\inst{1} \and
Roberto Lotufo\inst{1} \and
Jayr Pereira\inst{1,2}}
\authorrunning{Domingos Júnior et al.}
%
\institute{Universidade Estadual de Campinas (UNICAMP), Campinas--SP, Brazil \and
Universidade Federal do Cariri (UFCA), Juazeiro do Norte--CE, Brazil\\
\and National Center for State Courts (NCSC), Williamsburg, Virginia, United States\\
\email{jayr.pereira@ufca.edu.br}
} 
\maketitle              
\begin{abstract}
This paper presents \textbf{BR-TaxQA-R}, a novel dataset designed to support question answering with references in the context of Brazilian personal income tax law. The dataset contains 715 questions from the 2024 official Q\&A document published by Brazil's Internal Revenue Service, enriched with statutory norms and administrative rulings from the \textit{Conselho Administrativo de Recursos Fiscais} (CARF). We implement a Retrieval-Augmented Generation (RAG) pipeline using OpenAI embeddings for searching and GPT-4o-mini for answer generation. We compare different text segmentation strategies and benchmark our system against commercial tools such as ChatGPT and Perplexity.ai using RAGAS-based metrics. Results show that our custom RAG pipeline outperforms commercial systems in \textit{Response Relevancy}, indicating stronger alignment with user queries, while commercial models achieve higher scores in \textit{Factual Correctness} and \textit{fluency}. These findings highlight a trade-off between legally grounded generation and linguistic fluency. Crucially, we argue that human expert evaluation remains essential to ensure the legal validity of AI-generated answers in high-stakes domains such as taxation. BR-TaxQA-R is publicly available at \url{https://huggingface.co/datasets/unicamp-dl/BR-TaxQA-R}.
\end{abstract}

\section{Introduction}

A longstanding challenge many independent judiciaries and administrative courts face is that, as the population grows, so does the number of cases, placing increasing pressure on courts and often exceeding their capacity~\cite{lai2024large}. The integration of Artificial Intelligence (AI), particularly through Natural Language Processing (NLP) techniques, offers the potential to significantly enhance judicial efficiency and effectiveness. This vision is exemplified by the \textquote{Smart Courts} initiative outlined in China’s Artificial Intelligence Development Plan, published by the State Council~\cite{wang2023intelligent}. AI has been applied to various legal domain tasks, ranging from Named Entity Recognition to Ruling Prediction with the overarching goal of improving judicial productivity~\cite{siqueira2024ulysses}.
Developments in other judicial systems suggest a growing interest in understanding how such technologies might contribute to improved access and operational capacity. For instance, the 2023 Year-End Report on the Federal Judiciary in the United States highlights both the opportunities and limitations of AI in the courtroom, noting its potential to assist litigants with limited resources and to streamline certain processes, while also cautioning against overreliance due to risks such as hallucinated content, data privacy concerns, and the challenges of replicating nuanced human judgment\footnote{\url{https://www.supremecourt.gov/publicinfo/year-end/2023year-endreport.pdf}}.

The successful application of NLP techniques in the legal domain relies on the availability of specialized datasets. There is an ongoing effort to narrow the resource gap for training and evaluating NLP systems in Brazilian Portuguese, considering the variety of legal tasks involved. For example, \cite{siqueira2024ulysses} developed a large corpus for the Brazilian legal domain, proposing a methodology to extract and preprocess legal documents from the judiciary, legislative, and executive branches of government. This corpus is aimed at being used for pretraining tasks, but still requires further processing for downstream applications. \cite{da2024datasets} proposed semantic similarity datasets based on published decisions from two appeals bodies of Brazilian Federal and Administrative Courts, creating a unique resource for jurisprudence and case law research. \cite{vitorio2024building} constructed a human-annotated, relevance feedback dataset for legal information retrieval based on legislative documents from the Brazilian Congress. Although focused on a specific scenario, relevance feedback datasets are a crucial step toward developing robust legal question answering systems. Finally, \cite{presa2024evaluating} leveraged the curated Tax Law Question Answering (QA) manual for corporate entities, published by Brazil's Internal Revenue Service (i.e. \textit{Receita Federal do Brasil} or RFB), to create a QA dataset that includes legal document references supporting the answers. This dataset was used to evaluate LLMs' ability to generate answers when provided with the gold passage as context, enabling the analysis of both answer correctness and faithfulness to the supporting references.

In this paper, we propose BR-TaxQA-R, a dataset for Brazilian Tax Law Question Answering with supporting references focused on personal income tax law. BR-TaxQA-R extends the work of~\cite{presa2024evaluating} by enabling the evaluation of the complete QA pipeline, incorporating all legal document references, both explicitly and implicitly cited in the answers. As additional supporting context, BR-TaxQA-R includes a curated set of rulings from CARF, the administrative appeals court handling federal tax disputes in Brazil. These rulings compose the case law portion of the dataset, providing real-world interpretations and applications of Personal Income Tax regulations. BR-TaxQA-R enables the evaluation of complete Retrieval-Augmented Generation (RAG) pipelines, encompassing both the information retrieval and answer generation stages, and provides a baseline for future research. Our evaluation indicates that simple sliding-window segmentation achieves good results, and that incorporating relevant jurisprudence further improves performance. Although closed-source commercial tools employing LLM-based search pipelines achieve superior performance, retrieval in the legal domain remains a challenging task.

The key contributions of this paper are:
\begin{itemize}
    \item We introduce BR-TaxQA-R, a novel dataset for tax-related QA in Brazilian Portuguese, combining statutory and case law.
    \item We implement and benchmark a legal-domain-specific RAG pipeline using hierarchical segmentation and legal prompting.
    \item We evaluate the system against commercial LLM tools and discuss the trade-offs between legal traceability and linguistic fluency.
\end{itemize}

The remainder of this paper is organized as follows. Section~\ref{sec:methodology} details the methodology adopted to create the dataset, including the parsing of original questions and answers~(\ref{sec:question_extraction}), the acquisition of supporting legal documents~(\ref{sec:tax_regulations}), and the construction of the additional jurisprudence set~(\ref{sec:case_law_collection}). Section~\ref{sec:the-dataset} presents the published format and statistics of the BR-TaxQA-R dataset. Sections~\ref{sec:experiments} and~\ref{sec:results} describe the experiments conducted to evaluate the dataset and discuss the results. Finally, Section~\ref{sec:conclusions} concludes the paper.









\section{Dataset Acquisition Methodology} \label{sec:methodology}

This study aims to develop a dataset that can be used to train and evaluate a Retrieval-Augmented Generation (RAG) system~\cite{lewis2020rag} for answering questions related to Brazilian personal income tax law. The dataset construction followed a three-step methodology: (1) extraction of questions and answers from the official 2024 \textquote{Questions and Answers} document published by the RFB (cf. Section \ref{sec:question_extraction}), (2) collection and processing of tax regulations cited as references in the answers (cf. Section \ref{sec:tax_regulations}), and (3) retrieval of relevant administrative rulings from CARF to provide jurisprudential support (cf. Section \ref{sec:case_law_collection}). These components were combined to create a legally grounded and contextually rich dataset aligned with real-world tax guidance.


\subsection{Questions extraction} \label{sec:question_extraction}

The first step in the dataset acquisition process involved extracting the questions and answers from the official document \textquote{Questions and Answers} published by RFB for the year 2024~\cite{receitafederal2024irpf}. That document is available in PDF format, and we applied a combination of automated tools and manual verification to ensure accurate extraction.


Our approach was to extract as much information as possible from the document, preserving the original to allow further error correction and processing. In addition to the question and its answer, all legal document references provided to support the answers are relevant, and identifying and processing them represented most of this extraction work.

After extracting the text information from the PDF using a Python Library\footnote{https://pymupdf.readthedocs.io}, the document text was processed in two stages. The first stage consisted of splitting the text into the following parts: question, answer, legal documents supporting the answer, and links to other questions. The second stage consisted of processing the answer body to extract additional legal document references supporting the answer and additional links to other questions.

Since throughout the different questions the same legal document was referenced using different notation --- abbreviations or acronyms --- we applied a semi-automated document deduplication strategy using LLM support: we clustered the documents by their name's initial letter and passed the list to the LLM instructing for identifying and removing document's duplication, whenever the same part (e.g. article, paragraph, clauses) was referred to. The final list was manually verified to fix the remaining duplicates.

\subsection{Tax regulations} \label{sec:tax_regulations}

Tax regulations were obtained through a curated selection, defined by the references listed in the processed  \textquote{Questions and Answers} 2024 document. The original documents were retrieved as PDF files or HTML pages, ensuring the most up-to-date versions were selected. For PDF files, an automated download was performed, followed by text extraction using Python libraries and stored as text files.

In the case of HTML documents, web scraping techniques were applied to parse and clean the page content, removing the amended or revoked parts, respectively indicated by the \texttt{<strike>} (strikethrough text) and \texttt{<del>} (removed text) tags. The resulting text was saved as plain text files, named and organized according to the regulation identifiers. 


\subsection{Case law collection} \label{sec:case_law_collection}

Administrative rulings (case law) were collected through automated web scraping from the official repository of the Brazilian Ministry of Finance~\footnote{https://acordaos.economia.gov.br/solr/acordaos2/browse/}, the federal agency that houses CARF, and converted into plain text. Only 2023 rulings were processed, with previous years potentially added in a future dataset release.

To ensure that the inclusion of case law would genuinely enhance the relevance and contextual alignment of answers within BR-TaxQA-R, we adopted two primary qualitative criteria for selecting rulings through web scraping. First, we based the selection on the presence of keywords directly extracted from the questions in the \textquote{Questions and Answers} document. This constraint helped ensure that the retrieved decisions addressed legal issues analogous to those in the official guidance, thus avoiding the inclusion of unrelated jurisprudence.

Second, we applied a temporal filter to guarantee that the rulings reflected current legal interpretations. Only rulings published within one year of the 2024 \textquote{Questions and Answers} edition were considered. This time-bounded selection criterion aimed to mitigate the risk of referencing outdated precedents that might no longer align with current tax practices or administrative guidance. This aligns with retrospective studies, in which closed cases are indexed by the \textquote{date of death} and analyzed for their legal characteristics\cite{okamoto_metodologia_2022}. Likewise, selecting relevant rulings relies on subjective representations of legal concepts, which must be explicitly described and theoretically justified.

\section{BR-TaxQA-R} \label{sec:the-dataset}





We named the dataset BR-TaxQA-R, which stands for Brazilian Tax Question Answering with References. The dataset is composed of three main components: the \textbf{questions set}, the \textbf{sources set}, and the \textbf{case law set}.


\subsection{Questions set}

The question set contains 715 questions and answers extracted from the official document published by RFB. 117 ($\sim$16\%) out of the 715 questions do not reference any external documents, since their answers are not directly defined in a legal document; those questions were kept in the dataset for completion. The answers to several questions reference other questions within the document, and those links were captured. The question set was structured to hold the original data as much as possible, along with the scraped information:
\begin{itemize}
    \item \textbf{question\_number}: The question number, starting with 1, as referred to in the original document.
    \item \textbf{question\_summary}: A very brief description of the question subject, extracted from the original document.
    \item \textbf{question\_text}: The question itself, as originally posed.
    \item \textbf{answer}: Answer, as extracted from the original document. It is a list of strings, respecting the PDF formatting. It contains all the information provided after the question\_text and before a link to the document index, provided at the end of all questions.
    \item \textbf{answer\_cleaned}: The answer field after removing all explicit external references --- the legal documents captured in the sources set --- and all explicit inter-question references. External references were provided in the original document: explicitly, through grayed boxes, and implicitly, embedded in the answer text.
    \item \textbf{references}: The list of external references explicitly provided.
    \item \textbf{linked\_questions}: List of other questions linked in the provided answer.
    \item \textbf{formatted\_references}: The explicit external references, LLM-processed to separate the document title, articles, sections, paragraphs, and other specific parts mentioned.
    \item \textbf{embedded\_references}: External references are implicitly provided, embedded in the answer text.
    \item \textbf{formatted\_embedded\_references}: The implicit external references LLM-processed to separate the specific information mentioned, similar to the formatted\_references field.
    \item \textbf{all\_formated\_references}: Merge of formatted\_references and formatted\_embedded\_references fields, combining the information of the legal documents, and including the name of the text file (the file sub-field) containing each particular legal document has been captured in the dataset.
\end{itemize}

\subsection{Sources and case law sets}

The sources and case law sets compose a corpus supporting the answers provided to the questions set. The sources set contains all the legal documents listed as official sources for the answers provided in the original ``Questions and Answers'' document, and corresponds to the minimal legal documents set required for a RAG system to properly answer all the questions. 
The case law set contains CARF administrative rulings on the topics covered by the questions, which can potentially offer concrete examples on the concepts covered by the legal documents, and provide assistance to assertive answers by the same RAG system. 
Both the sources and case law sets have the following format:
\begin{itemize}
    \item \textbf{filename}: The legal document scraped data filename, as referred to within the all\_formated\_references field in the questions set.
    \item \textbf{filedata}: The scraped legal document information, extracted as text data.
\end{itemize}

\subsection{Dataset statistics}



Although the dataset is relatively small, the legal domain introduces significant complexity when selecting relevant segments from the supporting documents for answering questions. The case law documents can improve answer quality, but they also increase the overall pipeline complexity, as they are numerous and vary significantly in structure. Table \ref{tab:dataset_statistics} summarizes the dataset size.

Tables \ref{tab:dataset_links_statistics} and \ref{tab:most_referred_sources} present statistics on the number of links found in the answers to other questions and external documents. While there is one question with 20 links to others, most answers do not reference other questions, suggesting they are mainly independent. The number of external links per question is more heterogeneous, which can be interpreted as an indicator of complexity: it is reasonable to assume that the 25 questions with more than 10 external references are more challenging for a Q\&A system to answer correctly. Among the 478 external documents, only 10 account for over half of all references. Given the significant variation in length, these documents also present additional challenges for the information retrieval stage.

\begin{table}[t]
\centering
\caption{BR-TaxQA-R size statistics.}
\label{tab:dataset_statistics}
\scriptsize
\begin{tabular}{
>{\raggedleft\arraybackslash}p{1.75cm}
>{\raggedleft\arraybackslash}p{1.3cm}
>{\raggedleft\arraybackslash}p{1.4cm}
>{\raggedleft\arraybackslash}p{1.7cm}
>{\raggedleft\arraybackslash}p{1.7cm}
}
\toprule
 & \textbf{Questions} & \textbf{Answers} & \textbf{Source documents} & \textbf{Case-Law documents} \\ 
\midrule
count           & 715       &  715      &    478    &  7204 \\
min words       &   3       &    6      &     24    &   425 \\
max words       &  74       & 3118      & 165830    & 75584 \\
mean words      &  19.11    &  143.36   &   4546.46 &  3171.70 \\
median words    &  17.00    &   81.00   &    649.00 &  1983.50 \\
\bottomrule
\end{tabular}
\end{table}

\begin{table}[t]
\centering
\caption{BR-TaxQA-R question links statistics.}
\label{tab:dataset_links_statistics}
\scriptsize
\begin{tabular}{
>{\raggedleft\arraybackslash}p{1.5cm}
>{\centering\arraybackslash}p{1.8cm}
>{\centering\arraybackslash}p{1.8cm}
>{\centering\arraybackslash}p{1.8cm}
>{\centering\arraybackslash}p{1.8cm}
}
\toprule
 & \textbf{Links to other questions} & \textbf{Explicit external links} & \textbf{Implicit external links} & \textbf{Total external links} \\ 
\midrule
without   & 478 (66.85\%) &  151 (21.12\%)    &  524 (73.29\%)   &  117 (16.36\%)   \\
>2 and <10 & 51 (7.13\%) & 248 (34.69\%) & 55 (7.69\%) & 287 (40.14\%) \\
$\geq10$ & 4 (0.56\%) & 5 (0.70\%) & 3 (0.42\%) & 25 (3.50\%) \\
minimum   &   0   &    0    &    0    &    0    \\
maximum   &  20   &   18    &   16    &   21    \\
mean      &  0.69 &    2.07 &    0.63 &    2.70 \\
median    &  0.00 &    2.00 &    0.00 &    2.0  \\
\midrule
question max & 129 & 442 and 560 & 177 & 177, 442, 560 \\
\bottomrule
\end{tabular}
\end{table}

\begin{table}[]
\centering
\caption{10 most referred source documents by BR-TaxQA-R questions.}
\label{tab:most_referred_sources}
\scriptsize
\begin{tabular}{
>{\raggedleft\arraybackslash}p{0.5cm}
>{\raggedleft\arraybackslash}p{1.3cm}
@{\hspace{0.3cm}}
>{\raggedright\arraybackslash}p{7.6cm}
>{\raggedleft\arraybackslash}p{0.8cm}
}
\toprule
& \textbf{Reference count} & \textbf{Document} & \textbf{Word count}\\ 
\midrule
& 284 &  Decreto nº 9.580, de 22 de novembro de 2018 & 165830 \\
& 123 &  Instrução Normativa RFB nº 1500, de 29 de outubro de 2014 & 28269 \\
&  68 &  Lei nº 9.250, de 26 de dezembro de 1995 & 6452 \\
&  63 &  Instrução Normativa SRF nº 83, de 11 de outubro de 2001 & 3177 \\
&  52 &  Instrução Normativa SRF nº 208, de 27 de setembro de 2002 & 10118 \\
&  49 &  Lei nº 7.713, de 22 de dezembro de 1988 & 6636 \\
&  45 &  Instrução Normativa RFB nº 2178, de 05 de março de 2024 & 3997 \\
&  44 &  Instrução Normativa SRF nº 84, de 11 de outubro de 2001 & 4539 \\
&  37 &  Instrução Normativa RFB nº 1585, de 31 de agosto de 2015 & 26504 \\
&  22 &  Instrução Normativa SRF nº 81, de 11 de outubro de 2001 & 4880 \\
\midrule
Total & 787 & 53.32\% (1476) \\
\bottomrule
\end{tabular}
\end{table}

\section{Experiments} \label{sec:experiments}

This section describes the experiments conducted using BR-TaxQA-R. The experiments were designed to evaluate the performance of a custom Retrieval-Augmented Generation (RAG) system, which was implemented using the BR-TaxQA-R dataset source files as the knowledge base. We also assess the performance of two commercial tools, ChatGPT and Perplexity.ai, using the same set of questions from the BR-TaxQA-R dataset. The experiments aim to establish a baseline for the RAG system's performance in answering tax-related questions, comparing different segmentation strategies, and evaluating the results against commercial tools. In the next subsections, we describe the custom RAG system, the commercial tools used for comparison, and the evaluation metrics employed to assess the performance of the generated answers.

\subsection{Custom RAG system}

We implemented a custom RAG system using the BR-TaxQA-R dataset as the knowledge base. The system is designed to efficiently retrieve relevant information from the sources and case law sets, generate accurate answers to user queries, and provide explicit references to the legal documents used in the answer generation process. Following the principles outlined in~\cite{lewis2020rag}, the RAG system is structured into three main components: data preparation, indexing, and answer generation, ensuring that retrieved content is seamlessly integrated into the response generation pipeline.

\subsubsection{Data Preparation}

The data preparation includes the document segmentation and indexing for the Information Retrieval RAG stage.
We considered 2 data segmentation approaches for the \textbf{sources} and \textbf{case law} datasets, each one making increasing usage of the documents' internal structure:

\begin{itemize}
    \item \textbf{Sliding-window}, considering 2048-token windows and 1024-token stride, producing regular-sized overlapping segments.

    \item \textbf{Langchain Recursive Character Text Splitter}\footnote{https://python.langchain.com/docs/how\_to/recursive\_text\_splitter/}, which is recommended for generic text, splitting the text recursively according to a given separators list, until the resulting segments are small enough. In its default configuration, the provided separators try to keep paragraph contents in a single chunk, using very little information about the text's internal structure. We provided a customized separators list, including the documents containing statutory law hierarchy with many splitting points; the expected effect is that the recursive splitter would break the segments at those separators occurrences as much as possible, resulting in segments with meaningful boundaries according to the documents' original internal hierarchical structure. We considered chunks up to 1000-character length, with at most 100-character overlap.

\end{itemize}



\subsubsection{Indexing}

We adopted the dense passage retrieval approach~\cite{karpukhin2020dense}: once the documents were segmented, we indexed them using the \texttt{text-embedding-3-small} commercial model offered through OpenAI API~\footnote{https://openai.com/index/new-embedding-models-and-api-updates/}. This model is designed to generate dense vector representations of text, which can be used for similarity-based retrieval. The embeddings were generated for each of the two segmentation strategies and saved using FAISS (Facebook AI Similarity Search) \cite{douze2025faisslibrary} to enable efficient similarity-based retrieval. For each segmentation variant, a separate FAISS index was created. The FAISS \texttt{IndexFlatL2} type was employed, which computes the L2 (Euclidean) distance for nearest neighbor searches.
This step was repeated for each segmentation approach applied to the BR-TaxQA-R dataset. For case law documents, only the sliding-window segmentation was considered, due to the lack of hierarchical structure.

We implemented a retrieval function that uses the FAISS index to retrieve relevant chunks based on user queries. The retrieval process involves embedding the user query using the same \texttt{text-embedding-3-small} model and querying the FAISS index to find the \texttt{top k} most similar document segments.

\subsubsection{Answer Generation}

The context retrieved by FAISS is fed into a prompt-based RAG system powered by OpenAI’s \texttt{gpt-4o-mini}. The prompt is meticulously crafted to emulate the behavior of a virtual assistant specializing in Brazilian tax law. To enhance interpretability and improve legal reasoning, the system employs Chain-of-Thought (CoT) prompting, guiding the model to articulate intermediate reasoning steps before producing the final answer. This helps ensure that conclusions are logically grounded in the retrieved legal text. The final version of the prompt was designed to ensure that responses:

\begin{itemize}
    \item Are derived solely from the provided context;
    \item Contain no direct references to the context or user interaction;
    \item Include citations of applicable legal sources (norms and articles only) at the end of the response in a structured list;
    \item Avoid mid-response citations or references to document structure such as paragraphs or subitems;
    \item Follow naming conventions (e.g., ``Decreto Nº 9.580'' instead of ``RIR/2018'').
\end{itemize}

If not enough information is found in the retrieved context, the model is instructed to return a fallback answer indicating the system is still learning.
This structured response format, consisting of a generated answer and a list of cited legal sources, enables consistent and automated evaluation of the RAG pipeline across different segmentation strategies.



\subsection{Commercial Tools} 

In addition to evaluating our proposed RAG system, an assessment was conducted using prominent commercial Large Language Models (LLMs) equipped with integrated web search or deep search capabilities. The tools examined included:

\begin{itemize}
    \item \textbf{ChatGPT} (utilizing GPT-4o and GPT-4o mini models), with its search integration.
    \item \textbf{Perplexity AI}, employing its Deep Research feature~\footnote{https://www.perplexity.ai/hub/blog/introducing-perplexity-deep-research}.
    \item \textbf{Grok 3}, leveraging its DeepSearch functionality.
\end{itemize}

The primary objective of this comparative analysis was to benchmark the performance of these state-of-the-art commercial systems. We aimed to determine whether their search-augmented responses could effectively approximate the accuracy and completeness of the ground-truth answers. This evaluation sought to understand the capabilities and limitations of readily available, market-leading generative AI tools in retrieving factual information and providing valid, traceable sources for complex Brazilian tax inquiries, using the same question set applied to our custom RAG model.

\subsection{Evaluation Metrics}  

To quantitatively assess the performance of the responses generated by the commercial tools detailed in the previous section, a dedicated evaluation framework was employed, comparing its outputs against the established ground truth. This framework is built upon the RAGAS (Retrieval-Augmented Generation Assessment) library \cite{es-etal-2024-ragas}, a specialized Python package suited for evaluating generated text against reference text.

A custom Python script automated this evaluation. It processed a dataset structured in JSON format, containing the original question (\texttt{question\_text}), the ground truth answer (\texttt{answer}), and the candidate response generated by the commercial tool (\texttt{candidate}). The RAGAS library, within this script, utilized Langchain wrappers to interface with specified LLMs (e.g., \texttt{gpt-4o-mini}) and embedding models (e.g., \texttt{text-embedding-3-small}) for calculating certain evaluation metrics, as described below:

\begin{itemize}
    \item \textbf{Response Relevancy}: Measures how relevant the answer is to the original question by using an LLM to generate alternative questions from the answer's content, then calculating the average cosine similarity between their embeddings and the original question. Higher scores reflect better alignment with the query intent, penalizing incomplete or redundant answers.
    \item \textbf{Factual Correctness}: Refers to the degree to  which the candidate answer aligns with verified knowledge or ground truth claims, as if how much it was entailed by them. Factuality is measured by comparing model completions against trusted datasets or external knowledge bases. The metric is computed by matching generated answers to reference answers and checking how much data as statements remained, was lost, and created.
    \item \textbf{Semantic Similarity}: Semantic similarity will measure how closely a model's output matches a text used as reference. It will help detect whether responses preserve intended information. The metric is typically computed using sentence embeddings, and Higher scores indicate closer semantic alignment between the candidate and the reference. Factual Correctness may use it to match referred claims.
    \item \textbf{BLEU Score}: A precision-based metric that evaluates how many character sequences in particular order from the candidate answer can be overlapped to the reference texts. Scores range from 0 to 1, where higher scores indicate better overlap. BLEU is sensitive to exact word matches and word order, as it was firstly formulated to evaluate translated text \cite{papineni2002bleu}.
    \item \textbf{ROUGE Score}: A set of metrics that compares model output to reference summaries based on sensitivity (which is the fraction of correctly selected data from all relevant entailments) of n-grams, longest common subsequences, and skip-bigram matches. ROUGE-L is a widely known variant that captures fluency and structure through sequence alignment \cite{lin-2004-rouge}.
\end{itemize}

\section{Results} \label{sec:results}

We evaluated the performance of our custom RAG system using multiple segmentation strategies and compared it against leading commercial tools. Table~\ref{tab:results} presents a summary of the results across all evaluation metrics.

\begin{table}[]
\centering
\caption{Evaluation metrics for answer generation across different systems. All the Custom RAG used OpenAI's \texttt{gpt-4o-mini} for generating responses.}
\label{tab:results}
\scriptsize
\begin{tabular}{@{}p{4.9cm}
@{\hspace{-0.1cm}}
>{\raggedleft\arraybackslash}p{1.45cm}
>{\raggedleft\arraybackslash}p{1.2cm}
>{\centering\arraybackslash}p{1.8cm}
>{\raggedleft\arraybackslash}p{0.8cm}
@{\hspace{0.2cm}}
>{\centering\arraybackslash}p{1.6cm}@{}}
\toprule 
\textbf{Method} & \textbf{Resp. Relevancy} & \textbf{Factual Corr.} & \textbf{Semantic Similarity} & \textbf{BLEU} & \textbf{ROUGE-L} \\ 
\midrule
Custom RAG Systems \\
\midrule
Recursive segmentation                  & 0.791 & 0.286 & 0.765 & 0.185 & 0.241 \\
Sliding-window segmentation             & 0.819 & 0.313 & 0.766 & 0.178 & 0.248 \\
Recursive segmentation + case law       & 0.811 & 0.296 & 0.763 & 0.175 & 0.241 \\
Sliding-window segmentation + case law  & \textbf{0.829} & 0.327 & 0.768 & \textbf{0.190} & 0.248 \\
Only case law w/ sliding-window        & 0.818 & 0.209 & 0.744 & 0.149 & 0.207 \\
\midrule
Commercial Tools \\
\midrule
ChatGPT + Search tool                   & 0.738 & 0.389 & \textbf{0.793} & 0.158 & \textbf{0.251} \\
Perplexity.ai + Deep Research           & 0.665 & \textbf{0.469} & 0.757 & 0.075 & 0.106 \\
Grok 3 + DeepSearch                     &  0.509 & 0.454 & 0.745 & 0.099 & 0.089 \\

\bottomrule
\end{tabular}
\end{table}

Among the custom RAG configurations, the \textbf{sliding-window segmentation with case law} achieved the highest score in \textit{Response Relevancy} (0.829), outperforming all other systems, including commercial tools. This suggests that retrieving overlapping segments enriched with administrative rulings contributes positively to aligning model outputs with user intent.

In terms of \textit{Factual Correctness}, commercial models performed better: \textbf{Perplexity.ai} reached the highest score (0.469), followed by \textbf{ChatGPT} (0.389). Despite using grounded references, our RAG system was outperformed in this dimension, likely due to the broader training data and advanced retrieval mechanisms available to commercial tools.

The \textbf{sliding-window + case law} configuration also performed competitively in \textit{BLEU} (0.190) and \textit{ROUGE-L} (0.248), indicating that its generated answers were structurally and lexically similar to the gold standard. Nevertheless, \textbf{ChatGPT} led in \textit{Semantic Similarity} (0.793) and attained the best \textit{ROUGE-L} score (0.251), reflecting superior fluency and semantic alignment.


The \textbf{case law only} configuration yielded an interesting result: although it performed worse across most metrics, it achieved the second-highest \textit{Response Relevancy} score. This outcome suggests that relying solely on jurisprudential content (without normative references) limits the model’s ability to generate precise and legally grounded responses: the answers might be relevant, but not easily verifiable against the corresponding legislation.

Although commercial tools achieved better results, there remains room for improvement, as demonstrated in the literature where RAGAS \textit{Factual Correctness} and \textit{Semantic Similarity} metrics have been shown to align well with human evaluations~\cite{roychowdhury2024evaluation}, \cite{oro2024evaluating}. The verified performance on these metrics reinforces the understanding that retrieval remains a challenging task in the legal domain~\cite{magesh2405hallucination}, \cite{paul2024legal}, \cite{feng2024legal}. Furthermore, the application of jurisprudence can help bridge the gap between the abstract concepts encapsulated in legal statutes and regulations and the real-world facts described in user questions~\cite{xiao2023legal}, \cite{su2024stard}.

Overall, our results reveal a trade-off between legal traceability and linguistic fluency:
\begin{itemize}
    \item Our domain-specific RAG pipeline excels in \textit{relevance} and \textit{contextual precision};
    \item Commercial tools generate more fluent and complete answers, but often lack explicit grounding in legal sources.
\end{itemize}

\begin{figure}[h]
    \centering
    \includegraphics[width=\textwidth]{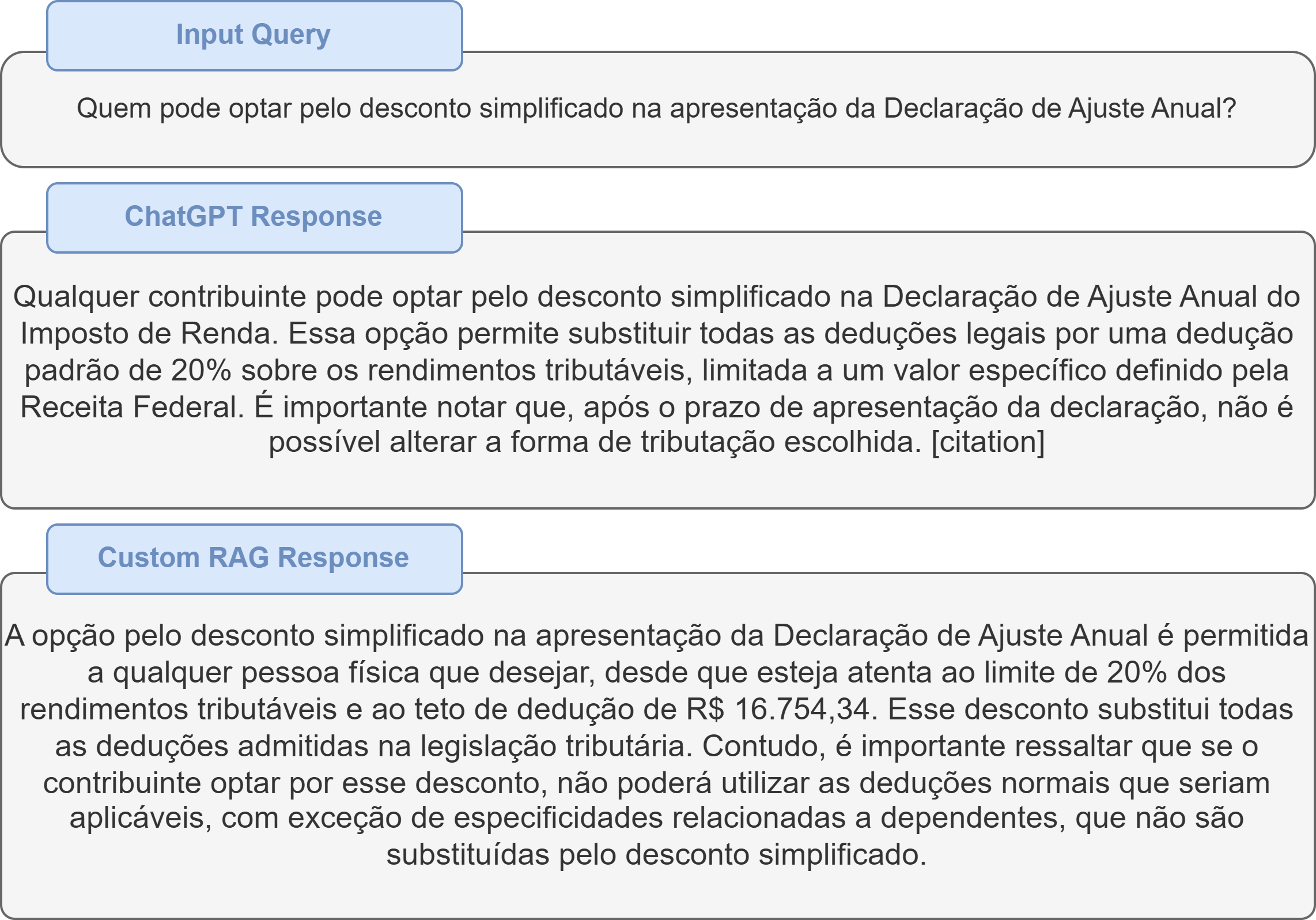}
    \caption{Illustration of the trade-off between contextual precision and linguistic fluency.}

    \label{fig:chatgpt_vs_custom}
\end{figure}

Figure \ref{fig:chatgpt_vs_custom} illustrates the trade-off between contextual precision and fluency. We consider the following example: \textquote{Who can opt for the standard deduction in the annual tax filing declaration?}. This case demonstrates the contrasting behavior of the two models:

\begin{itemize}

    \item \textbf{Custom RAG}: The response from the domain-specific RAG model provides a concise and legally precise description, directly reflecting the relevant tax regulations. It lists the exact criteria required for opting for the simplified deduction, aligning closely with the formal language and structure typical of regulatory documents. However, this approach tends to prioritize accuracy over readability, resulting in a less conversational tone.
    \item \textbf{ChatGPT}: In contrast, the ChatGPT response adopts a more conversational style, using natural language that is generally easier to read. It captures the main points effectively but lacks the precise legal references found in the RAG response. This broader, more accessible phrasing can be advantageous for non-specialist audiences but risks omitting critical legal nuances.

\end{itemize}

This example highlights the broader trend observed in our experiments: domain-specific models excel in precise, contextually accurate responses, while general-purpose commercial tools like ChatGPT often favor fluency and comprehensiveness at the expense of explicit legal traceability.

These findings underscore the importance of human expert evaluation in legal question answering. High scores in metrics such as \textit{semantic similarity} or \textit{ROUGE} do not guarantee legal adequacy. In several cases, fluent answers generated by commercial tools were factually incorrect or unsupported by authoritative legal documents--a critical issue in high-stakes contexts like tax guidance.

\section{Conclusions} \label{sec:conclusions}

This study introduced BR-TaxQA-R, a novel dataset designed to support the development and evaluation of Retrieval-Augmented Generation (RAG) systems in the domain of Brazilian personal income tax law. By combining statutory documents, administrative rulings (CARF decisions), and an extensive set of official questions and answers published by RFB, the dataset provides a valuable resource for both academic research and applied legal NLP.


Our experiments demonstrated that a custom RAG pipeline, carefully tailored to the legal domain through legal-specific prompting and employing simple sliding-window segmentation over the legal corpus, achieved strong performance in terms of \textit{Response Relevancy}, particularly when jurisprudence on the question topics was available. However, commercial systems such as ChatGPT, which benefit from broader training data and advanced retrieval mechanisms, outperformed our model in \textit{Factual Correctness} and fluency.

These findings suggest that while specialized systems can be more focused and legally grounded, they may still fall short in naturalness and completeness compared to state-of-the-art general-purpose tools.
Importantly, the evaluation results also emphasize the need for human assessment in legal QA tasks. Metrics such as semantic similarity or BLEU/ROUGE alone are insufficient to guarantee that an answer is legally valid or practically useful. In our case, some high-scoring answers from the RAG system lacked critical legal nuance, while ChatGPT occasionally provided fluent but ungrounded content. Thus, expert evaluation remains essential to ensure the legal accuracy and trustworthiness of AI-generated responses.

Future work includes incorporating multi-year CARF decisions and improving response calibration mechanisms. We also aim to refine human-in-the-loop evaluation protocols to better capture legal adequacy, traceability, and user trust in automated systems.

% ...


%
%
%
\bibliographystyle{splncs04}
\bibliography{references}
\end{document}